\documentclass[runningheads,a4paper]{llncs}

\usepackage{amssymb,amsmath}
\usepackage[caption=false]{subfig}
\usepackage{graphicx}
\usepackage{color}

\usepackage{url}
\urldef{\mailsa}\path|{ambra.demontis,paolo.russu,battista.biggio,fumera,roli}@diee.unica.it|

\newcommand{\ie}{\emph{i.e.}}
\newcommand{\eg}{\emph{e.g.}}
\newcommand{\etal}{\emph{et al.}}

\newcommand{\cf}{\emph{cf.}}

\newcommand{\vct}[1]{\ensuremath{\boldsymbol{#1}}} 

\newcommand{\set}[1]{\ensuremath{\mathcal{#1}}}
\newcommand{\con}[1]{\ensuremath{\mathsf{#1}}}

\newcommand{\T}{\ensuremath{\top}}

\newcommand{\argmin}{\operatornamewithlimits{\arg\,\min}}

\newcommand{\myparagraph}[1]{\smallskip \noindent \textbf{#1.}}

\begin{document}

\mainmatter  

\title{On Security and Sparsity of Linear Classifiers for Adversarial Settings}

\titlerunning{On Security and Sparsity of Linear Classifiers for Adversarial Settings}

\author{Ambra Demontis
\and Paolo Russu\and Battista Biggio\and\\
 Giorgio Fumera\and Fabio Roli}
\authorrunning{A. Demontis, P. Russu, B. Biggio, G. Fumera and F. Roli}

\institute{Department of Electrical and Electronic Engineering,\\
University of Cagliari, Piazza d'Armi 09123, Cagliari, Italy\\
\mailsa}

\toctitle{On Security and Sparsity of Linear Classifiers for Adversarial Settings}
\tocauthor{A. Demontis, P. Russu, B. Biggio, G. Fumera and F. Roli}
\maketitle

\newcommand{\oc}{$8$-\emph{gon}}

\begin{abstract}
Machine-learning techniques are widely used in security-related applications, like spam and malware detection.
However, in such settings, they have been shown to be vulnerable to \emph{adversarial} attacks, including the deliberate manipulation of data at test time to evade detection.
In this work, we focus on the vulnerability of linear classifiers to evasion attacks. 
This can be considered a relevant problem, as linear classifiers have been increasingly used in embedded systems and mobile devices for their low processing time and memory requirements. 
We exploit recent findings in robust optimization to investigate the link between regularization and security of linear classifiers, depending on the type of attack. 
We also analyze the relationship between the sparsity of feature weights, which is desirable for reducing processing cost, and the security of linear classifiers. We further propose a novel octagonal regularizer that allows us to achieve a proper trade-off between them.
Finally, we empirically show how this regularizer can improve classifier security and sparsity in real-world application examples including spam and malware detection.
\end{abstract}

\section{Introduction}
\label{sect:introduction}

Machine-learning techniques are becoming an essential tool in several application fields such as marketing, economy and medicine.
They are increasingly being used also in security-related applications, like spam and malware detection, despite their vulnerability to \emph{adversarial} attacks, \ie, the \emph{deliberate} manipulation of training or test data, to subvert their operation; \eg, spam emails can be manipulated (at test time) to evade a trained anti-spam classifier~\cite{dalvi04,lowd05,lowd05-ceas,kolcz09,nelson08,barreno06-asiaccs,biggio10-ijmlc,biggio13-ecml,biggio14-tkde,biggio14-ijprai,huang11,zhang16-tcyb}. 

In this work, we focus on the security of linear classifiers. These classifiers are particularly suited to mobile and embedded systems, as the latter usually demand for strict constraints on storage, processing time and power consumption. Nonetheless, linear classifiers are also a preferred choice as they provide easier-to-interpret decisions (with respect to nonlinear classification methods).
For instance, the widely-used SpamAssassin anti-spam filter exploits a linear classifier~\cite{biggio10-ijmlc,nelson08}.\footnote{See also \url{http://spamassassin.apache.org}.}
Work in the adversarial machine learning literature has already investigated the security of linear classifiers to evasion attacks~\cite{kolcz09,biggio10-ijmlc}, suggesting the use of more evenly-distributed feature weights as a mean to improve their security. Such a solution is however based on heuristic criteria, and a clear understanding of the conditions under which it can be effective, or even optimal, is still lacking. 
Moreover, in mobile and embedded systems, \emph{sparse} weights are more desirable than evenly-distributed ones, in terms of processing time, memory requirements, and interpretability of decisions.

In this work, we shed some light on the security of linear classifiers, leveraging recent findings from~\cite{xu09,sra11,livni12} that highlight the relationship between classifier regularization and robust optimization problems in which the input data is potentially corrupted by noise (see Sect.~\ref{sect:background}).
This is particularly relevant in adversarial settings as the aforementioned ones, since evasion attacks can be essentially considered a form of noise affecting the non-manipulated, initial data (\eg, malicious code before obfuscation).
Connecting the work in~\cite{xu09,sra11,livni12} to adversarial machine learning aims to
help understanding what the optimal regularizer is against different kinds of adversarial \emph{noise} (attacks). We analyze the relationship between the sparsity of the weights of a linear classifier and its security in Sect.~\ref{sect:sec}, where we also propose an octagonal-norm regularizer to better tune the trade-off arising between sparsity and security.
In Sect.~\ref{sect:experiment}, we empirically evaluate our results on a handwritten digit recognition task, 
and on real-world application examples including spam filtering and detection of malicious software (malware) in PDF files. 
We conclude by discussing the main contributions and findings of our work in Sect.~\ref{sect:conclusions}, while also sketching some promising research directions. 

\section{Background}
\label{sect:background}

In this section, we summarize the attacker model previously proposed in~\cite{biggio13-ecml,biggio14-tkde,biggio14-ijprai,huang11}, and the link between regularization and robustness discussed in~\cite{xu09,sra11,livni12}.

\subsection{Attacker's Model}

To rigorously analyze possible attacks against machine learning and devise principled countermeasures, a formal model of the attacker has been proposed in~\cite{barreno06-asiaccs,biggio13-ecml,biggio14-tkde,biggio14-ijprai,huang11}, based on the definition of her goal (\eg, evading detection at test time), knowledge of the classifier, and capability of manipulating the input data.

\myparagraph{Attacker's Goal}
Among the possible goals, here we focus on evasion attacks, where the goal is to modify a single malicious sample (\eg, a spam email) to have it misclassified as legitimate (with the largest confidence) by the classifier~\cite{biggio13-ecml}.

\myparagraph{Attacker's Knowledge}
The attacker can have different levels of knowledge of the targeted classifier; she may  have \emph{limited} or \emph{perfect} knowledge about the training data, the feature set, and the classification algorithm~\cite{biggio13-ecml,biggio14-tkde}.
In this work, we focus on \emph{perfect-knowledge} (worst-case) attacks.

\myparagraph{Attacker's Capability}
In evasion attacks, the attacker is only able to modify malicious instances.
Modifying an instance usually has some cost. Moreover, arbitrary modifications to evade the classifier may be ineffective, if the resulting instance loses its malicious nature (\eg, excessive obfuscation of a spam email could make it unreadable for humans).
This can be formalized by an application-dependent constraint. 
As discussed in~\cite{wang14-icdm}, two kinds of constraints have been mostly used when modeling real-world adversarial settings, leading one to define \emph{sparse} ($\ell_{1}$) and \emph{dense} ($\ell_{2}$) attacks.
The \textbf{$\ell_{1}$-norm} yields typically a sparse attack, as it represents the case when the cost depends on the number of modified features. For instance, when instances correspond to text (\eg, the email's body) and each feature represents the occurrences of a given term in the text, the attacker usually aims to change as few words as possible.
The \textbf{$\ell_{2}$-norm} yields a dense attack, as it represents the case when the cost of modifying features is proportional to the distance between the original and modified sample in Euclidean space. For example, if instances are images, the attacker may prefer making small changes to many or even all pixels, rather than significantly modifying only few of them. This amounts to (slightly) blurring the image, instead of obtaining a salt-and-pepper noise effect (as the one produced by sparse attacks)~\cite{wang14-icdm}.

\myparagraph{Attack Strategy} It consists of the procedure for modifying samples, according to the attacker's goal, knowledge and capability, formalized as an optimization problem. 
Let us denote the legitimate and malicious class labels respectively with $-1$ and $+1$, and assume that the classifier's decision function is $f(\mathbf x) = {\rm sign} \left ( g(\mathbf x) \right )$, where $g(\mathbf x) = \vct w^{\T} \vct x + b \in \mathbb R$ is a linear discriminant function with feature weights $\vct w \in \mathbb R^{\con d}$ and bias $b \in \mathbb R$, and $\mathbf x$ is the representation of an instance in a $\con d$-dimensional feature space. 
Given a malicious sample $\vct x_0$, the goal is to find the sample $\vct x^*$ that minimizes the classifier's discriminant function $g(\cdot)$ (\ie, that is classified as legitimate with the highest possible confidence) subject to the constraint that $\vct x^*$ lies within a distance $d_{\rm max}$ from $\vct x_0$:
\begin{eqnarray}
\label{eq:ev1}
\vct x^* = &&  \argmin_{\vct x}  g(\vct x ) \\
\label{eq:ev2}
{\rm s. t.} && d(\vct x, \vct x_0) \leq d_{\rm max} \, ,
\end{eqnarray}
where the distance measure $d(\cdot,\cdot)$ is defined in terms of the cost of data manipulation (\eg, the number of modified words in each spam)~\cite{dalvi04,lowd05,biggio13-ecml,biggio14-tkde,zhang16-tcyb}. For sparse and dense attacks, $d(\cdot,\cdot)$ corresponds respectively to the $\ell_{1}$ and $\ell_{2}$ distance.

\subsection{Robustness and Regularization}
\label{sect:RobustnessAndRegularization}

The goal of this section is to clarify the connection between regularization and input data uncertainty, leveraging on the recent findings in~\cite{xu09,sra11,livni12}.
In particular, Xu~\etal{}~\cite{xu09} have considered the following \emph{robust} optimization problem: 
\begin{equation}
\label{eq:xu}
 \min_{\vct w,b} \; \max_{\vct u_1,..., \vct u_{\con m} \in \set U} \, \sum_{i=1}^{\con m} \left(1- y_i( \vct w^{\T} (\vct x_i -\vct u_i) + b)\right)_{+} \, ,
\end{equation}
where $\left( z \right)_{+}$ is equal to $z \in \mathbb R$ if $z > 0$ and $0$ otherwise, $\vct u_1,..., \vct u_{\con m} \in \set U$ define a set of bounded perturbations of the training data $\{\vct x_{i}, y_{i}\}_{i=1}^{\con m} \in \mathbb R^{\con m} \times \{-1,+1\}^{\con m}$, and the so-called \emph{uncertainty set} $\set U$ is defined as $\set U \overset{\Delta}{=} \left \{ (\vct u_1, \ldots , \vct u_{\con m}) | \sum_{i=1}^{\con m} \| \vct u_i \|^*  \leq c \right \}$, being $\| \cdot \|^*$ the dual norm of $\| \cdot \|$.
Typical examples of uncertainty sets according to the above definition include $\ell_{1}$ and $\ell_{2}$ balls~\cite{xu09,sra11}.

\begin{figure}[t]
	\centering
	\includegraphics[trim=20 10 20 5, clip, height=0.32\textwidth]{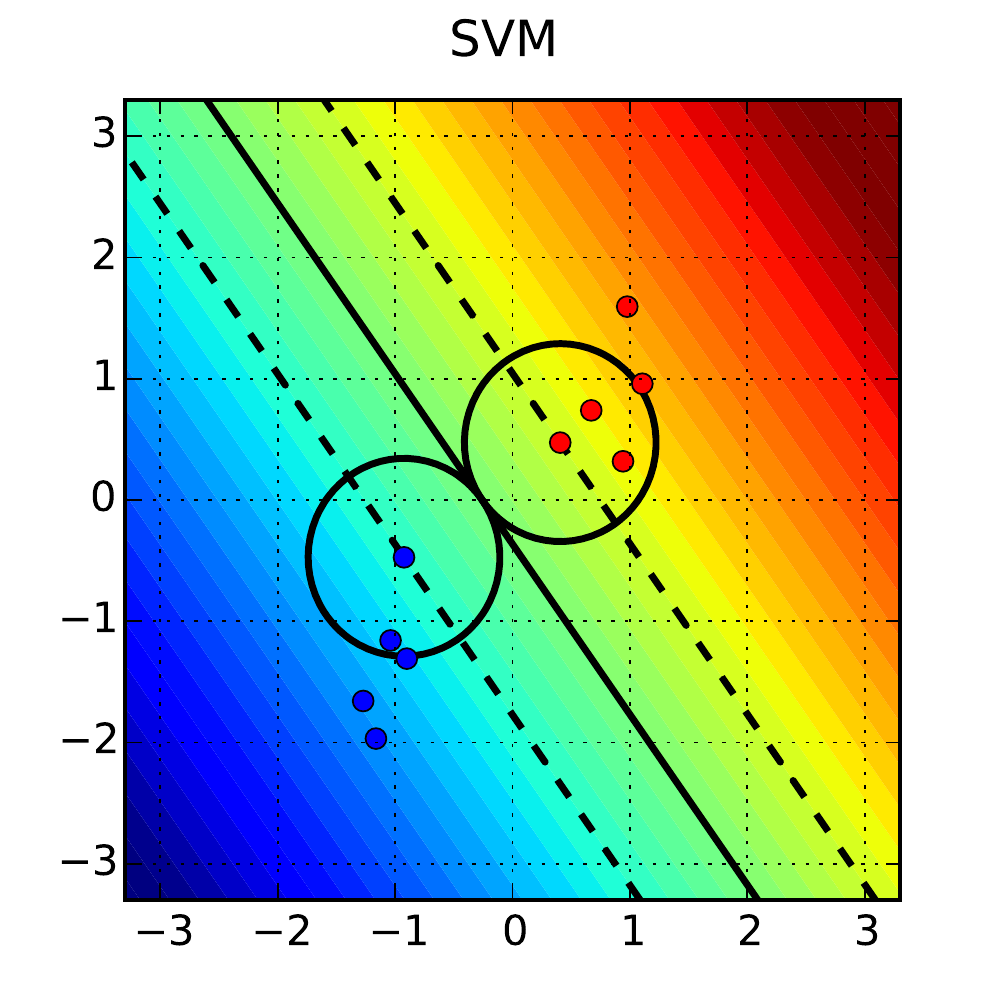}
	\includegraphics[trim=20 10 20 5, clip, height=0.32\textwidth]{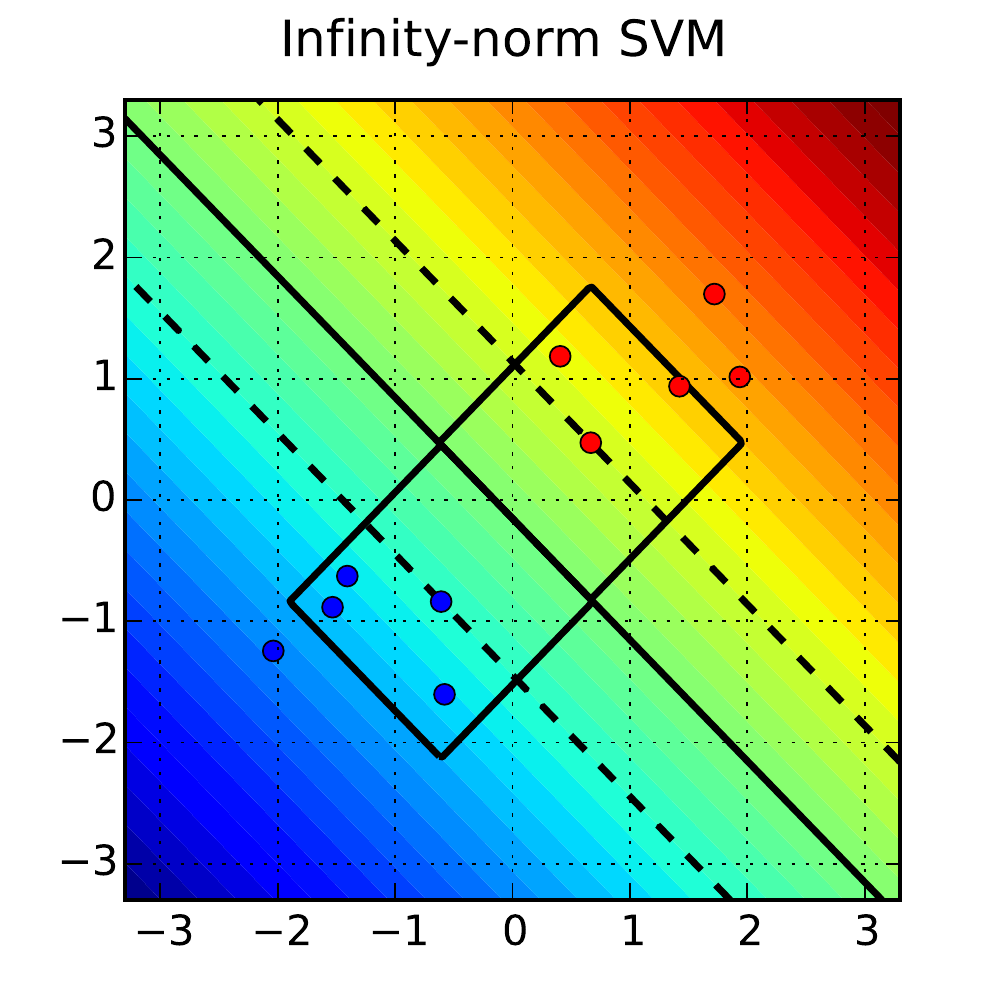}		
	\includegraphics[trim=20 10 20 5, clip, height=0.32\textwidth]{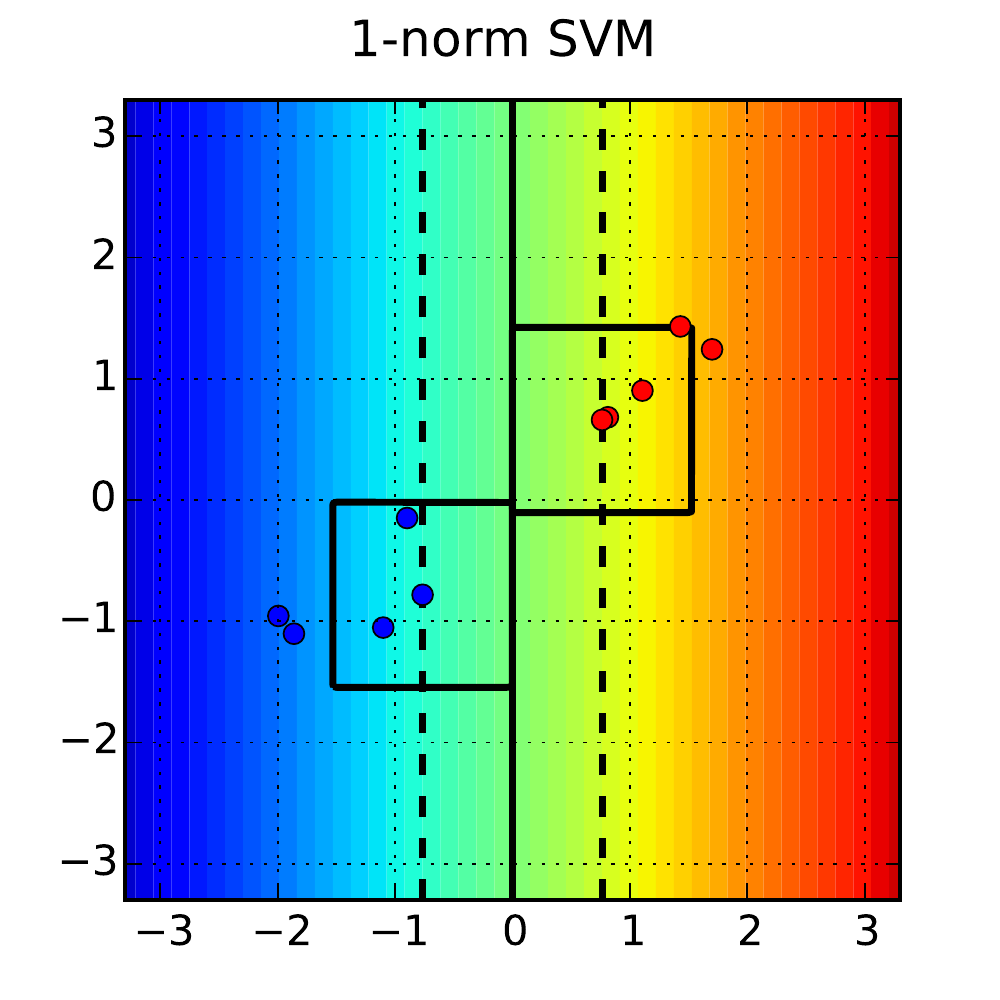}
	\includegraphics[trim=5 5 5 7, clip, height=0.32\textwidth]{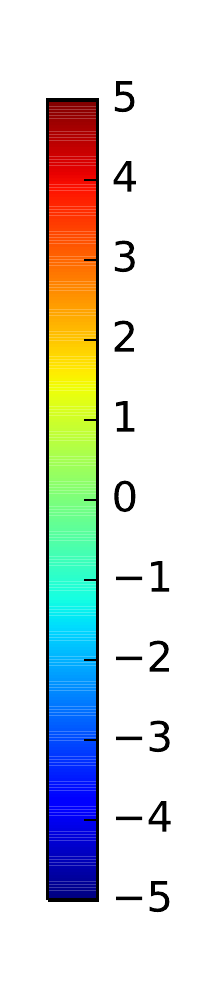}
	\caption{Discriminant function $g(\vct x)$ for SVM, Infinity-norm SVM, and 1-norm SVM (in colors). The decision boundary ($g(\vct x)=0$) and margins ($g(\vct x) =\pm 1$) are respectively shown with black solid and dashed lines. Uncertainty sets are drawn over the support vectors to show how they determine the orientation of the decision boundary.}
	\label{fig:dual_norm}
\end{figure}

Problem~\eqref{eq:xu} basically corresponds to minimizing the hinge loss for a two-class classification problem under worst-case, bounded perturbations of the training samples $\vct x_{i}$, \ie, a typical setting in robust optimization~\cite{xu09,sra11,livni12}.
Under some mild assumptions easily verified in practice (including non-separability of the training data), the authors have shown that the above problem is equivalent to the following non-robust, regularized optimization problem (\cf~Th.~3 in~\cite{xu09}):
\begin{eqnarray}
 \min_{\vct w,b} \;  c \| \vct w \| + \sum_{i=1}^{\con m} \left(1- y_i( \vct w^{\T} \vct x_i  + b)\right)_{+} \,.
\end{eqnarray}
This means that, if the $\ell_{2}$ norm is chosen as the dual norm characterizing the uncertainty set $\set U$, then $\vct w$ is regularized with the $\ell_{2}$ norm, and the above problem is equivalent to a standard Support Vector Machine (SVM)~\cite{vapnik95}.
If input data uncertainty is modeled with the $\ell_{1}$ norm, instead, the optimal regularizer would be the $\ell_{\infty}$ regularizer, and vice-versa.\footnote{Note that the $\ell_{1}$ norm is the dual norm of the $\ell_{\infty}$ norm, and vice-versa, while the $\ell_{2}$ norm is the dual norm of itself.}
This notion is clarified in Fig.~\ref{fig:dual_norm}, where we consider different norms to model input data uncertainty against the corresponding SVMs; \ie, the standard SVM~\cite{vapnik95}, the Infinity-norm SVM~\cite{bennett00} and the 1-norm SVM~\cite{zhu04-nips} against $\ell_{2}$, $\ell_{1}$ and $\ell_{\infty}$-norm uncertainty models, respectively.

\section{Security and Sparsity}
\label{sect:sec} 
We discuss here the main contributions of this work. The result discussed in the previous section, similar to that reported independently in~\cite{livni12}, helps  understanding the security properties of linear classifiers in adversarial settings, in terms of the relationship between security and sparsity.
In fact, what discussed in the previous section does not only confirm the intuition in~\cite{kolcz09,biggio10-ijmlc}, \ie, that more uniform feature weighting schemes should improve classifier security by enforcing the attacker to manipulate more feature values to evade detection.
The result in~\cite{xu09,sra11,livni12} also clarifies the meaning of \emph{uniformity} of the feature weights $\vct w$.
If one considers an $\ell_{1}$ (sparse) attacker, facing a higher cost when modifying more features, it turns out that the optimal regularizer is given by the $\ell_{\infty}$ norm of $\vct w$, which tends to yield more uniform weights. In particular, the solution provided by $\ell_{\infty}$ regularization (in the presence of a strongly-regularized classifier) tends to yield weights which, in absolute value, are all equal to a (small) maximum value. This also implies that $\ell_{\infty}$ regularization does not provide a \emph{sparse} solution.

For this reason we propose a novel \emph{octagonal} (8gon) regularizer,\footnote{Note that octagonal regularization has been previously proposed also in \cite{bondell08}. However, differently from our work, the authors have used a pairwise version of the infinity norm, for the purpose of selecting (correlated) groups of features.} given as a linear (convex) combination of $\ell_{1}$ and $\ell_{\infty}$ regularization: 
\begin{equation}
\| \vct w \|_{\rm 8gon} = (1-\rho) \| \vct w \|_{1} + \rho \| \vct w \|_{\infty} \,
\end{equation}
where $\rho \in (0,1)$ can be increased to trade sparsity for security.

Our work does not only aim to clarify the relationships among regularization, sparsity, and \emph{adversarial} noise. We also aim to quantitatively assess the aforementioned trade-off on real-world application examples, to evaluate whether and to what the extent the choice of a proper regularizer may have a significant impact in practice. 
Thus, besides proposing a new regularizer and shedding light on uniform feature weighting and classifier security, the other main contribution of the present work is the experimental analysis reported in the next section, in which we consider both dense ($\ell_{2}$) and sparse ($\ell_{1}$) attacks, and evaluate their impact on SVMs using different regularizers. We further analyze the weight distribution of each classifier to provide a better understanding on how sparsity is related to classifier security under the considered evasion attacks.  

\section{Experimental Analysis}
\label{sect:experiment}

We first consider dense and sparse attacks in the context of handwritten digit recognition, to visually demonstrate their blurring and salt-and-pepper effect on images.
We then consider two real-world application examples including spam and PDF malware detection, investigating the behavior of different regularization terms against (\emph{sparse}) evasion attacks.

\myparagraph{Handwritten Digit Classification}
To visually show how evasion attacks work, we perform sparse and dense attacks on the MNIST digit data~\cite{LeCun95}. Each image is represented by a vector of 784 features, corresponding to its gray-level pixel values.
As in \cite{biggio13-ecml}, we simulate an adversarial classification problem where the digits $8$ and $9$ correspond to the legitimate and malicious class, respectively.

\myparagraph{Spam Filtering}
This is a well-known application subject to adversarial attacks.
Most spam filters include an automatic text classifier that analyzes the email's body text. In the simplest case Boolean features are used, each representing the presence or absence of a given term.
For our experiments we use the TREC 2007 spam track data, consisting of about 25000 legitimate and 50000 spam emails~\cite{trec07}.
We extract a dictionary of terms (features) from the first 5000 emails (in chronological order) using the same parsing mechanism of SpamAssassin, and then select the 200 most discriminant features according to the information gain criterion~\cite{sebastiani02}.
We simulate a well-known (\emph{sparse}) evasion attack in which the attacker aims to modify only few terms. Adding or removing a term amounts to switching the value of the corresponding Boolean feature~\cite{lowd05-ceas,kolcz09,biggio13-ecml,biggio14-tkde,zhang16-tcyb}.

\myparagraph{PDF Malware Detection}
Another application that is often targeted by attackers is the detection of malware in PDF files.
A PDF file can host different kinds of contents, like Flash and JavaScript. Such third-party applications can be exploited by attacker to execute arbitrary operations.
We use a data set made up of about 5500 legitimate and 6000 malicious PDF files. We represent every file using the 114 features that are described in~\cite{maiorca12-mldm}. They consist of the number of occurrences of a predefined set of keywords, where every keyword represents an action performed by one of the objects that are contained into the PDF file (\eg, opening another document that is stored inside the file).
An attacker cannot trivially remove keywords from a PDF file without corrupting its functionality. 
Conversely, she can easily add new keywords by inserting new object's operations. For this reason, we simulate this attack by only considering feature increments (decrementing a feature value is not allowed). Accordingly, the most convenient strategy to mislead a malware detector (classifier) is thus to insert as many occurrences of a \emph{given} keyword as possible, which is a sparse attack.\footnote{Despite no upper bound on the number of injected keywords may be set, we set the maximum value for each keyword to the corresponding one observed during training.}

\begin{figure*}[t]%
    \centering
    \subfloat[$\ell_{2}$]{\includegraphics[trim=15 15 15 15, clip,width=0.2\textwidth]{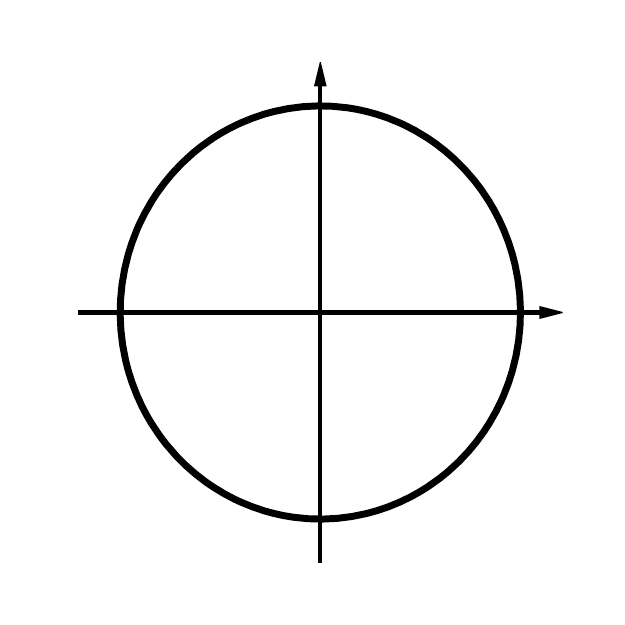}}
    \subfloat[$\ell_{\infty}$]{\includegraphics[trim=15 15 15 15, clip,width=0.2\textwidth]{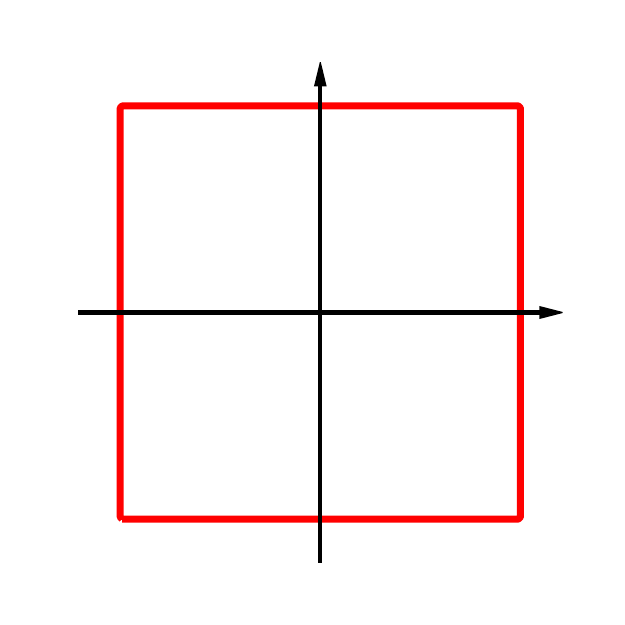}}
    \subfloat[$\ell_{1}$]{\includegraphics[trim=15 15 15 15, clip,width=0.2\textwidth]{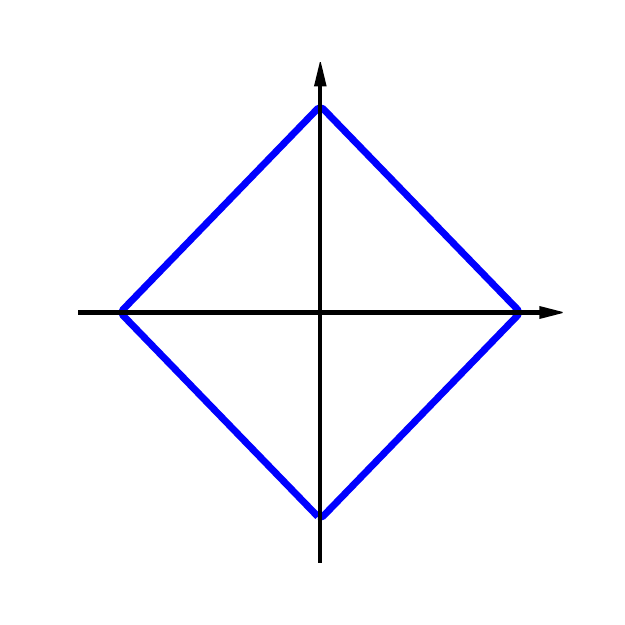}}
    \subfloat[elastic net]{\includegraphics[trim=15 15 15 15, clip,width=0.2\textwidth]{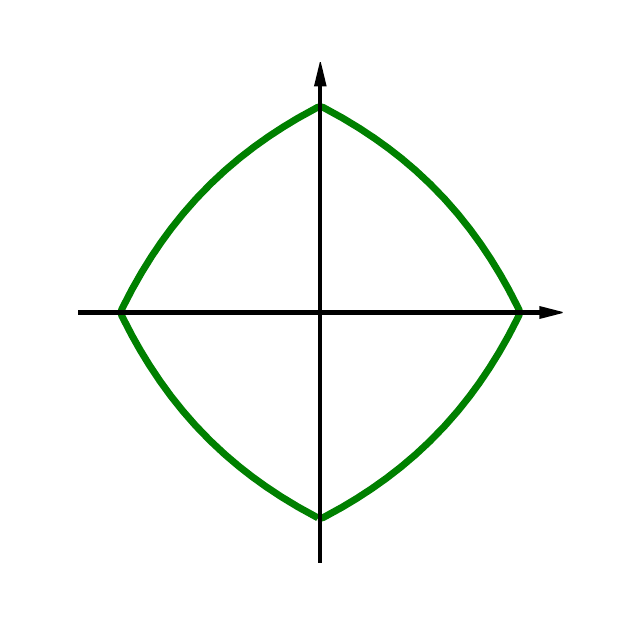}}
    \subfloat[octagonal]{\includegraphics[trim=15 15 15 15, clip,width=0.2\textwidth]{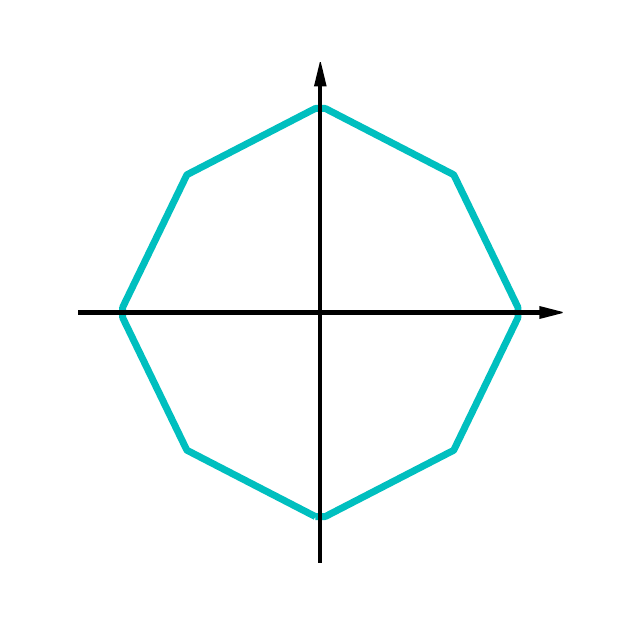}}
    \caption{Unit balls for different norms.}
    \label{fig:unitary_ball}
\end{figure*}

We consider different versions of the SVM classifier obtained by combining the hinge loss with the different regularizers shown in Fig.~\ref{fig:unitary_ball}.

\myparagraph{2-norm SVM (SVM)} This is the standard SVM learning algorithm~\cite{vapnik95}. It finds $\vct w$ and $b$ by solving the following quadratic programming problem:
\begin{eqnarray}
 \min_{\vct w,b} && \frac{1}{2}\| \vct w \|^{2}_{2} + C \sum_{i=1}^{\con m} \left( 1 - y_{i} g(\vct x_{i}) \right)_{+} \, ,
\end{eqnarray}
where $g(\vct x)=\vct w^{\T} \vct x + b$ denotes the SVM's linear discriminant function.
Note that $\ell_2$ regularization does not induce sparsity on $\vct w$.

\myparagraph{Infinity-norm SVM ($\infty$-norm)} In this case, the $\ell_{\infty}$ regularizer bounds the weights' maximum absolute value as $\| \vct w \|_{\infty} = \max_{j=1,\ldots,\con d} |w_{j}|$~\cite{bondell08}:
\begin{eqnarray}
 \min_{\vct w,b} && \| \vct w \|_{\infty} + C \sum_{i=1}^{\con m} \left( 1 - y_{i} g(\vct x_{i}) \right)_{+} \, .
\end{eqnarray}
As the standard SVM, this classifier is not sparse; however, the above learning problem can be solved using a simple linear programming approach.

\myparagraph{1-Norm SVM (1-norm)} Its learning algorithm is defined as~\cite{zhu04-nips}:
\begin{eqnarray}
\label{eq:1-norm}
 \min_{\vct w,b} && \| \vct w \|_{1} + C \sum_{i=1}^{\con m} \left( 1 - y_{i} g(\vct x_{i}) \right)_{+} \, .
\end{eqnarray}
The $\ell_{1}$ regularizer induces sparsity, while retaining convexity and linearity.

\myparagraph{Elastic-net SVM (el-net)} We use here the elastic-net regularizer~\cite{zou05-elasticnet}, combined with the hinge loss to obtain an SVM formulation with tunable sparsity:
\begin{eqnarray}
 \min_{\vct w,b} &&  (1-\lambda) \| \vct w \|_{1} + \frac{\lambda}{2} \| \vct w \|^2_{2} + C \sum_{i=1}^{\con m} \left( 1 - y_{i} g(\vct x_{i}) \right)_{+} \, .
\end{eqnarray}
The level of sparsity can be tuned through the trade-off parameter $\lambda \in (0,1)$.

\myparagraph{Octagonal-norm SVM (8gon)}   This novel SVM is based on our octagonal-norm regularizer, combined with the hinge loss: 
 \begin{eqnarray}
  \min_{\vct w,b} &&  (1-\rho) \| \vct w \|_{1} + \rho \| \vct w \|_{\infty}  + C \sum_{i=1}^{\con m} \left( 1 - y_{i} g(\vct x_{i}) \right)_{+} \, .
 \end{eqnarray}
The above optimization problem is linear, and can be solved using state-of-the-art solvers. The sparsity of $\vct w$ can be increased by decreasing the trade-off parameter $\rho \in (0,1)$, at the expense of classifier security.

\myparagraph{Sparsity and Security Measures} We evaluate the degree of sparsity $S$ of a given linear classifier as the fraction of its weights that are equal to zero:
\begin{equation}
\label{eq:S}
S = \frac{1}{\con d} |  \{ w_{j} | w_{j} = 0, j=1,\ldots,\con d  \}  | \, ,
\end{equation}
being $| \cdot |$ the cardinality of the set of null weights.

To evaluate security of linear classifiers, we define a measure $E$ of \emph{weight evenness}, similarly to~\cite{kolcz09,biggio10-ijmlc}, based on the ratio of the $\ell_{1}$ and $\ell_{\infty}$ norm:
\begin{eqnarray}
\label{eq:Sec}
E =  \frac{ \| \vct w \|_1 }{ \con d \| \vct w \|_{\infty}} \enspace ,
\end{eqnarray} 
where dividing by the number of features $\con d$ ensures that $E \in \left [\frac{1}{\con d},1\right]$, with higher values denoting more evenly-distributed feature weights. In particular, if only a weight is not zero, then $E=\frac{1}{\con d}$; conversely, when all weights are equal to the maximum (in absolute value), $E=1$.

\myparagraph{Experimental Setup} We randomly select 500 legitimate and 500 malicious samples from each dataset, and equally subdivide them to create a training and a test set. 
We optimize the regularization parameter $C$ of each SVM (along with $\lambda$ and $\rho$ for the Elastic-net and Octagonal SVMs, respectively) through 5-fold cross-validation,
maximizing the following objective on the training data:
\begin{equation}
{\rm AUC} + \alpha E+ \beta S 
\end{equation}
 where AUC is the area under the ROC curve, and $\alpha$ and $\beta$ are parameters defining the trade-off between security and sparsity.
We set $\alpha=\beta=0.1$ for the PDF and digit data, and $\alpha=0.2$ and $\beta=0.1$ for the spam data, to promote more secure solutions in the latter case. These parameters allow us to accept a marginal decrease in classifier security only if it corresponds to much sparser feature weights.
After classifier training, we perform evasion attacks on all malicious test samples, and evaluate the corresponding performance as a function of the number of features modified by the attacker. 
We repeat this procedure five times, and report the average results on the original and modified test data.

\begin{figure*}[t]
	\centering
	\includegraphics[width=0.495\textwidth]{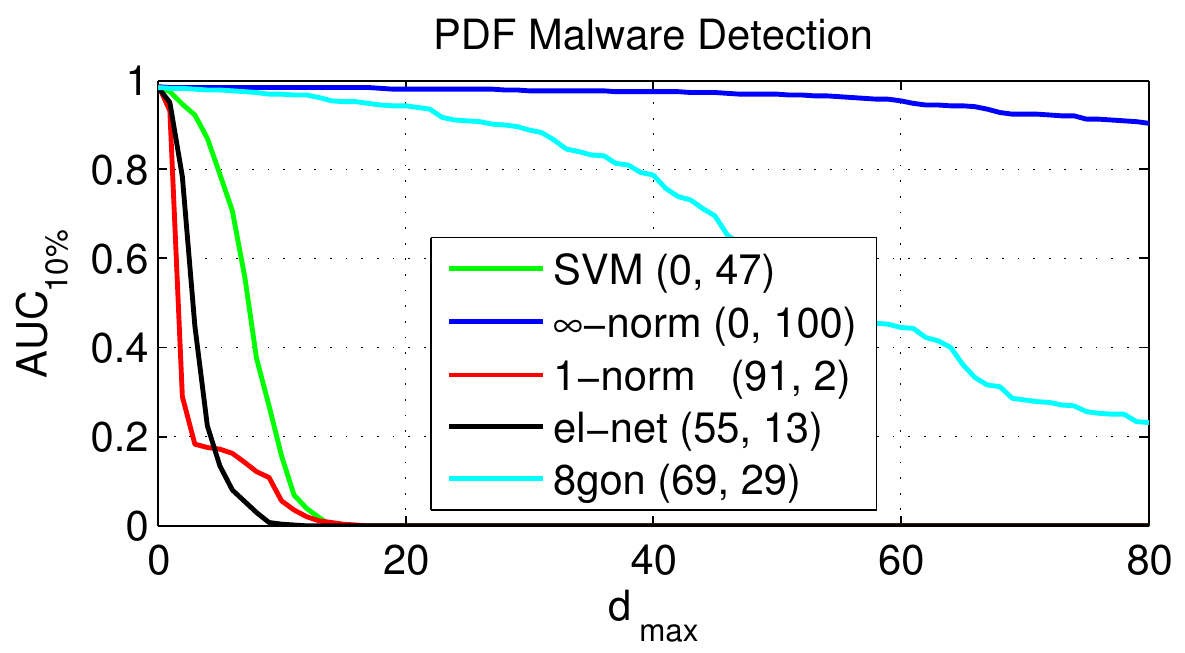}
	\includegraphics[width=0.495\textwidth]{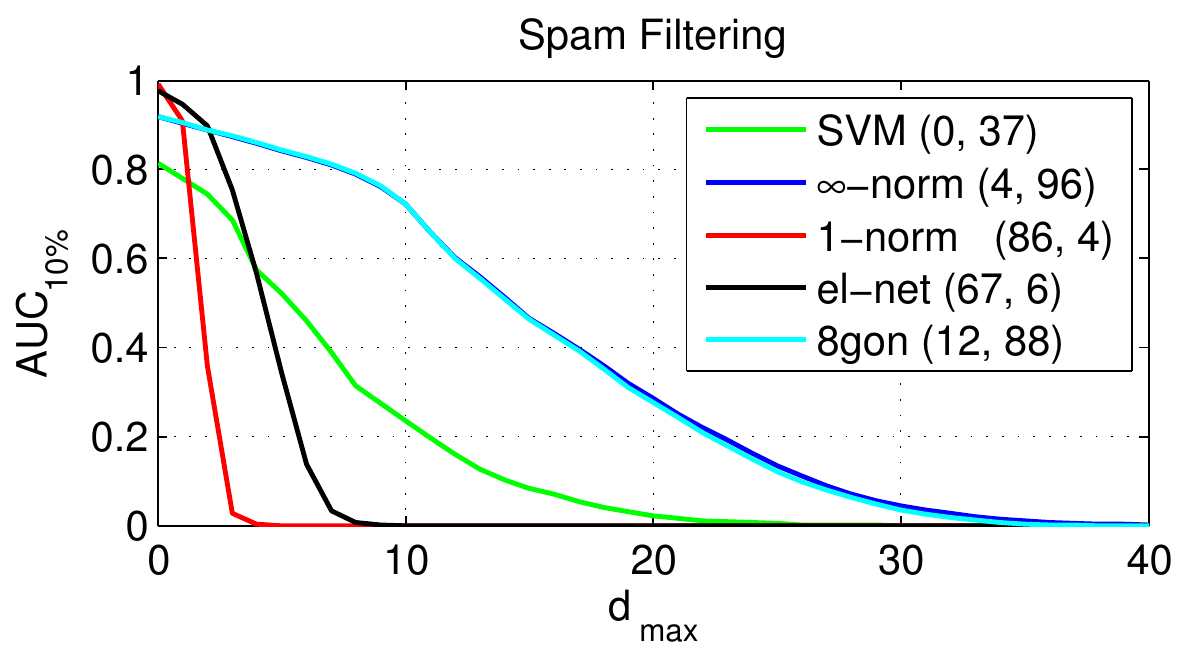}
	\caption{Classifier performance under attack for PDF malware and spam data, measured in terms of ${\rm AUC}_{10\%}$ against an increasing number $d_{\rm max}$ of modified features. For each classifier, we also report $(S,E)$ percentage values (Eqs.~\ref{eq:S}-\ref{eq:Sec}) in the legend.}
	\label{fig:real_app_under_attack}
\end{figure*}

\begin{figure*}[t]
\centering
	\includegraphics[width=0.19\textwidth]{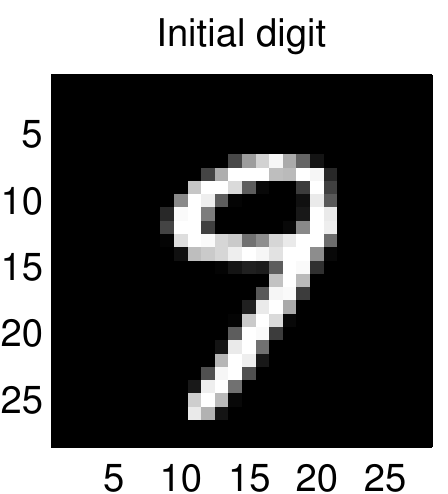} \\%
	\includegraphics[width=0.19\textwidth]{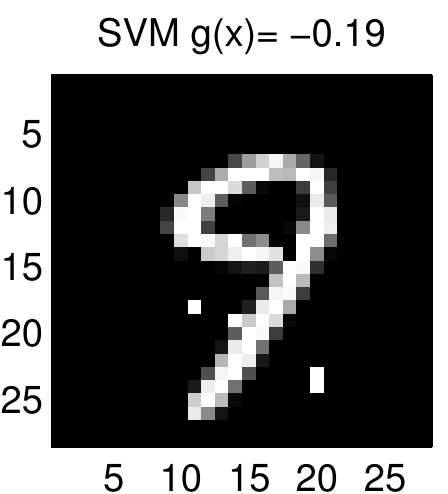}%
	\includegraphics[width=0.19\textwidth]{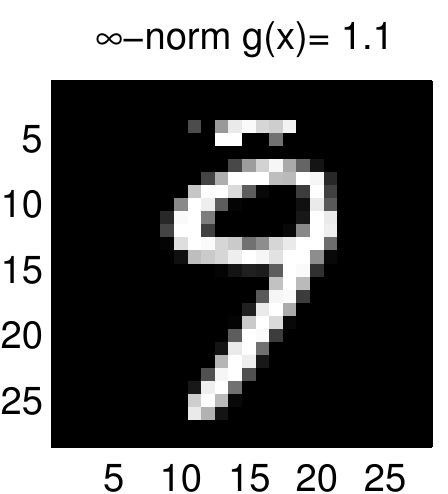}%
	\includegraphics[width=0.19\textwidth]{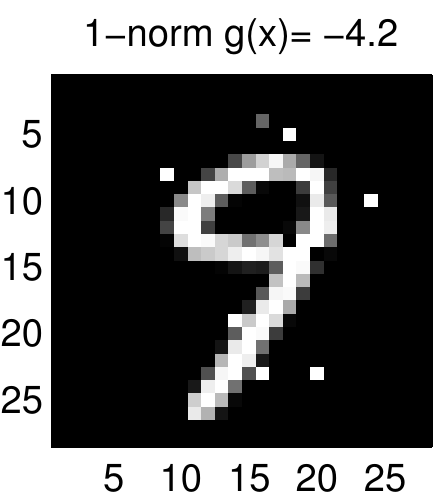}
	\includegraphics[width=0.19\textwidth]{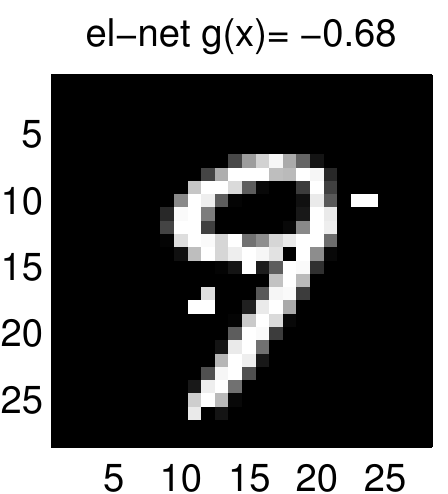}
	\includegraphics[width=0.19\textwidth]{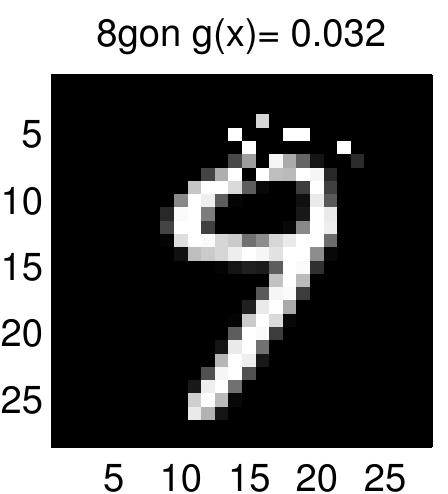} \\
	\includegraphics[width=0.19\textwidth]{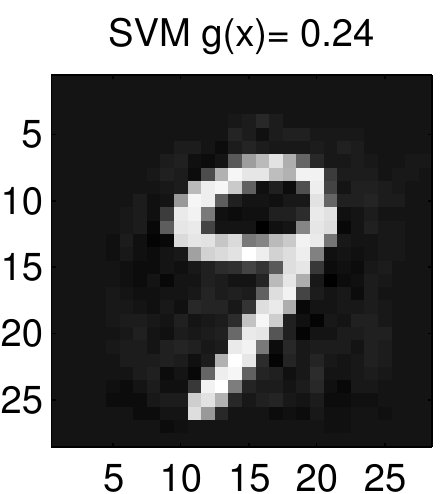}
	\includegraphics[width=0.19\textwidth]{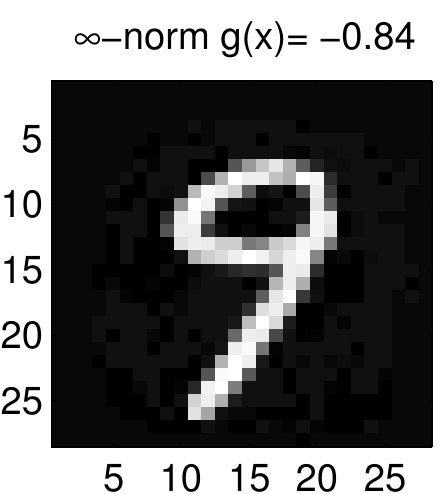}%
	\includegraphics[width=0.19\textwidth]{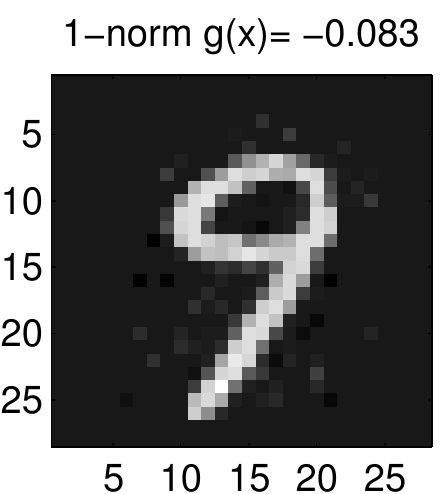}%
	\includegraphics[width=0.19\textwidth]{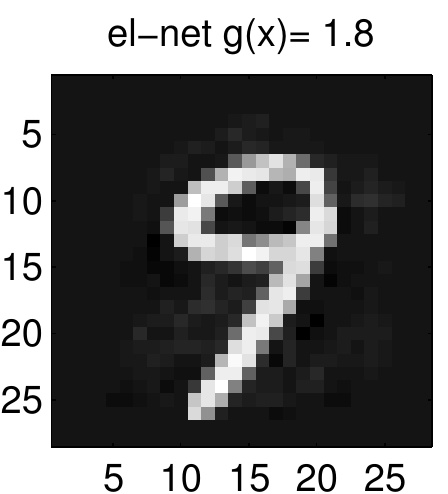}%
	\includegraphics[width=0.19\textwidth]{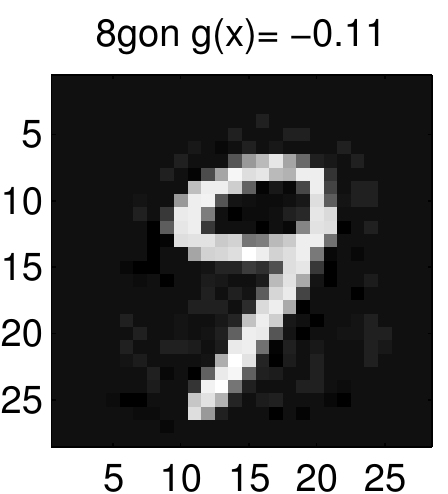}%
	\caption{Initial digit ``9'' (\emph{first row}) and its versions modified to be misclassified as ``8'' (\emph{second and third row}). Each column corresponds to a different classifier (from \emph{left} to \emph{right} in the second and third row): SVM, Infinity-norm SVM, 1-norm SVM, Elastic-net SVM, Octagonal SVM. \emph{Second row}: sparse attacks ($\ell_{1}$), with $d_{\rm max}=2000$. \emph{Third row}: dense attacks ($\ell_{2}$), with $d_{\rm max}=250$. Values of $g(\vct x) < 0$ denote a successful classifier evasion (\ie, more vulnerable classifiers).}
	\label{fig:digit_modified_samples}
\end{figure*}

\myparagraph{Experimental Results}
We consider first PDF malware and spam detection. In these applications, as mentioned before, only sparse evasion attacks make sense, as the attacker aims to minimize the number of modified features.
In Fig.~\ref{fig:real_app_under_attack}, we report the AUC at $10\%$ false positive rate for the considered classifiers, against an increasing number of words/keywords changed by the attacker.
This experiment shows that the most secure classifier under sparse evasion attacks is the Infinity-norm SVM, since its performance degrades more gracefully under attack. This is an expected result, given that, in this case, infinity-norm regularization corresponds to the dual norm of the attacker's cost/distance function. Notably, the Octagonal SVM yields reasonable security levels while achieving much sparser solutions, as expected (\cf{} the sparsity values $S$ in the legend of Fig.~\ref{fig:real_app_under_attack}). This experiment really clarifies how much the choice of a proper regularizer can be crucial in real-world adversarial applications.

By looking at the values reported in Fig.~\ref{fig:real_app_under_attack}, it may seem that the security measure $E$ does not properly characterize classifier security under attack; \eg, note how Octagonal SVM exhibits lower values of $E$ despite being more secure than SVM on the PDF data.
The underlying reason is that the attack implemented on the PDF data only considers feature increments, while $E$ generically considers any kind of manipulation. Accordingly, one should define alternative security measures depending on specific kinds of data manipulation.
However, the security measure $E$ allows us to properly tune the trade-off between security and sparsity also in this case and, thus, this issue may be considered negligible.

Finally, to visually demonstrate the effect of sparse and dense evasion attacks, we report some results on the MNIST handwritten digits. In Fig.~\ref{fig:digit_modified_samples}, we show the ``9'' digit image modified by the attacker to have it misclassified by the classifier as an ``8''.
These modified digits are obtained by solving Problem~\eqref{eq:ev1}-\eqref{eq:ev2} through a simple projected gradient-descent algorithm, as in \cite{biggio13-ecml}.\footnote{Note also that, in the case of sparse attacks, Problem~\eqref{eq:ev1}-\eqref{eq:ev2} amounts to minimizing a linear function subject to linear constraints. It can be thus formulated as a linear programming problem, and solved with state-of-the-art linear solvers.}
Note how dense attacks only produce a slightly-blurred effect on the image, while sparse attacks create more evident visual artifacts. 
By comparing the values of $g(\vct x)$ reported in Fig.~\ref{fig:digit_modified_samples}, one may also note that this simple example confirms again that Infinity-norm and Octagonal SVM are more secure against sparse attacks, while SVM and Elastic-net SVM are more secure against dense attacks.

\section{Conclusions and Future Work}
\label{sect:conclusions}
In this work we have shed light on the theoretical and practical implications of sparsity and security in linear classifiers. We have shown on real-world adversarial applications that the choice of a proper regularizer is crucial. In fact, in the presence of sparse attacks, Infinity-norm SVMs can drastically outperform the security of standard SVMs. We believe that this is an important result, as (standard) SVMs are widely used in security tasks without taking the risk of adversarial attacks too much into consideration. 
Moreover, we propose a new octagonal regularizer that enables trading sparsity for a marginal loss of security under sparse evasion attacks. This is extremely useful in applications where sparsity and computational efficiency at test time are crucial.
When dense attacks are instead deemed more likely, the standard SVM may be retained a good compromise. In that case, if sparsity is required, one may trade some level of security for sparsity using the Elastic-net SVM. 
Finally, we think that an interesting extension of our work may be to investigate the trade-off between sparsity and security also in the context of classifier poisoning (in which the attacker can contaminate the training data to mislead classifier learning)~\cite{biggio14-tkde,biggio12-icml,biggio15-icml}.

\end{document}